\pdfoutput=1
\documentclass[runningheads]{llncs}
\usepackage{epsfig}
\usepackage{graphicx}
\usepackage{amsmath,amssymb} 
\usepackage{color}
\usepackage{multirow}
\usepackage{makecell}
\usepackage{array}
\usepackage{subcaption}
\usepackage[misc]{ifsym}
\usepackage[pagebackref=true,breaklinks=true,colorlinks,bookmarks=false]{hyperref}
\begin{document}

\makeatletter
\newcommand{\printfnsymbol}[1]{%
  \textsuperscript{\@fnsymbol{#1}}%
}
\makeatother

\pagestyle{headings}
\mainmatter
\def\ECCVSubNumber{1479}  

\title{Semi-Siamese Training for Shallow Face Learning
} 

\titlerunning{Semi-Siamese Training for Shallow Face Learning}
\authorrunning{H. Du, et al.}
%

\author{Hang Du\inst{1,2}\thanks{Equal contribution. This work was performed at JD AI Research.}
\and Hailin Shi\inst{2}\printfnsymbol{1} 
\and Yuchi Liu\inst{2}
\and Jun Wang\inst{2} 
\and Zhen Lei\inst{3}
\and \\ Dan Zeng\inst{1}\textsuperscript{\Letter}
\and Tao Mei\inst{2} }
\institute{
Shanghai University, Shanghai, China. \\
\email{\{duhang, dzeng\}@shu.edu.cn} \\
\and
JD AI Research, Beijing, China.\\
\email{\{shihailin, wangjun492, tmei\}@jd.com, u6009551@anu.edu.au}\\ 
\and 
NLPR, Institute of Automation, Chinese Academy of Sciences, Beijing, China \\
\email{zlei@nlpr.ia.ac.cn}
}

\maketitle

\begin{abstract}

Most existing public face datasets, such as MS-Celeb-1M and VGGFace2, provide abundant information in both breadth (large number of IDs) and depth (sufficient number of samples) for training.
However, in many real-world scenarios of face recognition, the training dataset is limited in depth, \textit{i.e.} only two face images are available for each ID. 
\textit{We define this situation as Shallow Face Learning, and find it problematic with existing training methods.}
Unlike deep face data, the shallow face data lacks intra-class diversity. 
As such, it can lead to collapse of feature dimension and consequently the learned network can easily suffer from degeneration and over-fitting in the collapsed dimension. 
In this paper, we aim to address the problem by introducing a novel training method named Semi-Siamese Training (SST). 
A pair of Semi-Siamese networks constitute the forward propagation structure, and the training loss is computed with an updating gallery queue, conducting effective optimization on shallow training data. Our method is developed without extra-dependency, thus can be flexibly integrated with the existing loss functions and network architectures.
Extensive experiments on various benchmarks of face recognition show the proposed method significantly improves the training, not only in shallow face learning, but also for conventional deep face data.
\end{abstract}

\begin{figure}[t]
\centering
\includegraphics[scale=0.25]{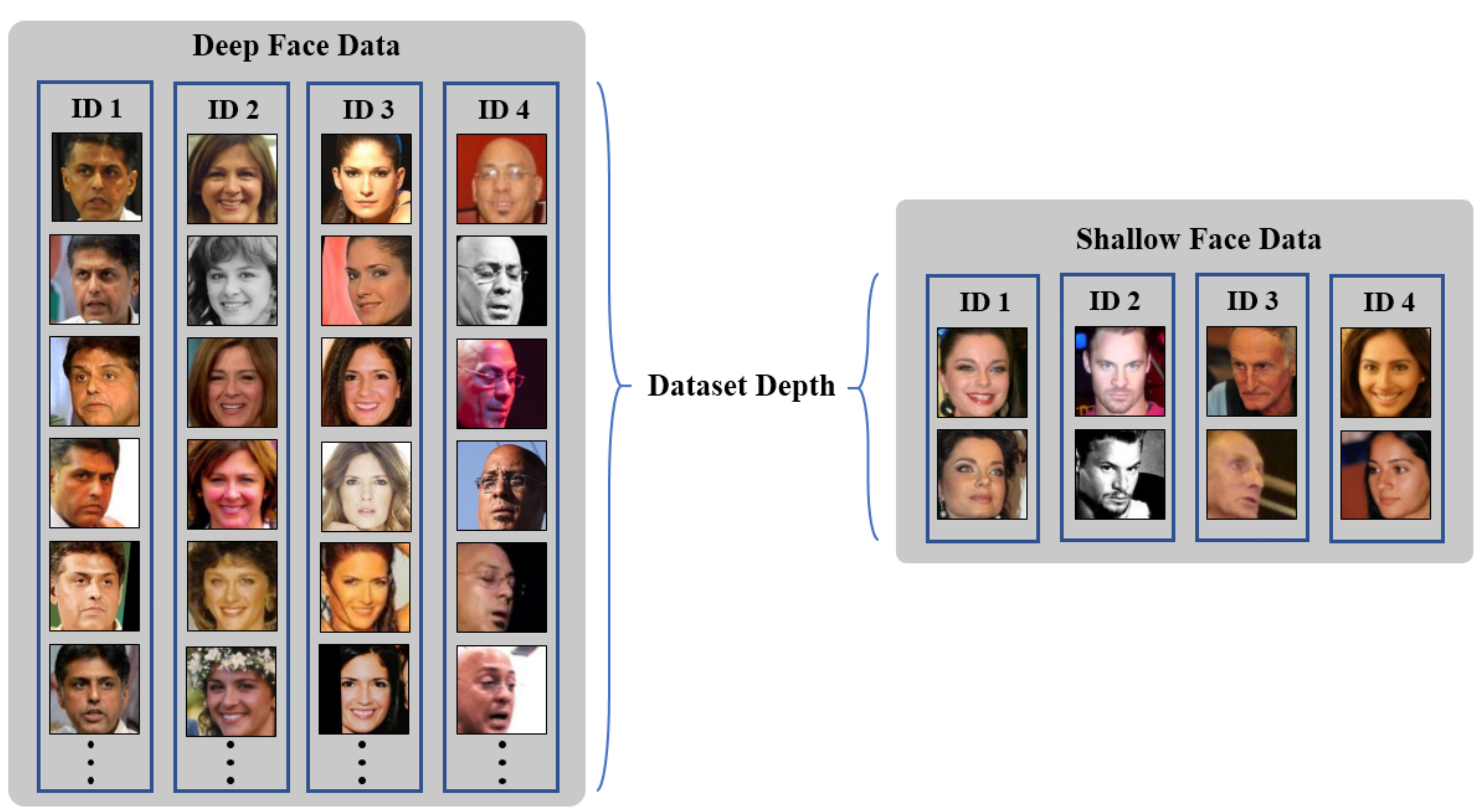}
\caption{Deep face data and shallow face data comparison in terms of data depth. Usually, only two images are available for each ID in shallow face data.}
\label{shallow_data}
\end{figure}

\section{Introduction}
Face Recognition (FR) has made remarkable advance and has been widely applied in the last few years.
It can be attributed to three aspects, including convolution neural networks (CNNs)~\cite{Simonyan2014Very,he2016deep,wang2017residual,hu2018squeeze}, loss functions~\cite{taigman2014deepface,sun2014deep,wen2016discriminative,ranjan2017l2,zhao2019regularface,wang2019mis} and large-scale training datasets~\cite{yi2014learning,guo2016ms,kemelmacher2016megaface,cao2018vggface2}. 
In recent years, the commonly used public training datasets, such as CASIA-WebFace~\cite{yi2014learning}, MS-Celeb-1M~\cite{guo2016ms} and VGGFace2~\cite{cao2018vggface2} \textit{etc.}, provide abundant information in not only breadth (large number of IDs), but also depth (dozens of face images for each ID). In this paper, we call this type of dataset as deep face data. Unfortunately, such deep face data is not available in many real-world scenarios. Usually, the training encounters the problem of ``shallow face data" in which only two face images are available for each ID (generally a registration photo and a spot photo, so-called ``gallery" and ``probe"). 
As a result, it lacks intra-class diversity, which prevents the network from effective optimization and leads to the collapse of feature dimension.
In such situation, we find the existing training methods suffer from either the model degeneration or the over-fitting issue.

In this paper, we regard the training on shallow face data as a particular task, named \textbf{Shallow Face Learning} (SFL). SFL is similar to the existing problem of Low-shot Learning (LSL)~\cite{fei2006one} in face recognition, but they have two significant differences. First, LSL performs close-set recognition~\cite{guo2017one,wu2017low,cheng2017know,wang2018feature}, while SFL includes open-set recognition task in which test IDs are excluded from training IDs.
Second, LSL requires pretraining in the source domain (with deep data) before finetuning to the target domain~\cite{zhu2019large,cheng2017know,yin2019feature}, however, the pretraining is not always a good choice for practical development of face recognition \textit{w.r.t} the following reasons: (1) the network architecture is fixed once the pretraining is done, thus it is inconvenient to change the architecture in the finetuning; (2) deploying new architectures needs restarting from the pretraining, while the pretraining is often time-consuming; (3) there exists domain gap between pretraining data and finetuning data, so the finetuning still suffers from the shallow data problem. Therefore, SFL argues to directly train from scratch on shallow face data. 

In brief, the objective of Shallow Face Learning is the effective training from scratch on shallow face data for open-set face recognition. 
We retrospect the current methods and study how they suffer from the shallow data problem.
In recent years, most of the prevailing deep face recognition methods~\cite{liu2017sphereface,wang2018cosface,wang2018additive,deng2019arcface,liu2019adaptiveface} are developed from the classification learning by softmax or its variants. They are built on a fully connected (FC) layer, the softmax function and the cross-entropy loss. The weights of the FC layer can be regarded as the prototypes which represent the center of each class. The learning objective is to maximize the prediction probability on the ground-truth class.
This routine shows great capability and efficiency to learn discrimination on deep data.
However, since the shallow data leads to the extreme lack of intra-class information, as shown in Section~\ref{problem}, we find this kind of training methods suffer from either model degeneration or over-fitting.

Another major routine in face recognition is the embedding learning methods~\cite{chopra2005learning,hadsell2006dimensionality,sun2014deep,schroff2015facenet,sohn2016improved}, which can learn face representation without the classification layer. For example, Contrastive loss~\cite{sun2014deep} and Triplet loss~\cite{schroff2015facenet} calculate pair-wise Euclidean distance and optimize the model over the sample relation. 
Generally, the embedding learning performs better than the classification learning when data becomes shallow.
The potential reason is that the embedding learning employs feature comparison between samples, instead of classifying them to the specific classes whose prototypes include large amount of parameters. 

However, the performance and efficiency of the embedding learning routine depends on the number of sample pairs matched batch-wisely, which is limited by the GPU memory and hard sampling strategy. 
In this paper, we desire to draw the advantage of embedding learning for achieving successful classification learning on shallow data. If we address the issues of model degeneration and over-fitting, the training can greatly benefit from the capability and efficiency of the classification learning. 
A straightforward solution comes up from the plain combination of the two routines, which employs sample features as the prototypes to initialize the FC weights, and runs classification learning with them. 
The similar modification on softmax has been suggested by the previous methods~\cite{zhu2019large}.
Specifically, for each ID of the shallow data, one photo is employed as the initial prototype, and the other photo is employed as training sample.
However, such prototype initialization brings still limited improvement when training on shallow data (\textit{e.g.} DP-softmax in Fig.~\ref{section_sss_with_various_losses}).
To explain this result, we assume that the prototype becomes too similar to its intra-class training sample, which leads to the extreme small gradient and impedes the optimization.

To overcome this issue, we propose to improve the training method from the perspective of enlarging intra-class diversity.
Taking Contrastive or Triplet loss as an example, the features are extracted by the backbone. The backbone can be regarded as a pair (or a triplet) of Siamese networks, since the parameters are fully shared between the networks. 
We find the crucial technique for the solution is to enforce the backbone being \textbf{Semi-Siamese}, which means the two networks have close (but not identical) parameters. One of the networks extracts the feature from gallery as the prototype, and the other network extracts the feature from probe as the training sample, for each ID in the training. The intra-class diversity between the features is guaranteed by the difference between the networks. 
There are many ways to constrain the two networks to have slight difference. For example, one can add a network constraint between their parameters during SGD (stochastic gradient descent) updating; or SGD updating for one, and moving-average updating for the other (like momentum proposed by~\cite{he2019momentum}). We conduct extensive experiments and find all of them contribute to the shallow face learning effectively. Furthermore, we incorporate the Semi-Siamese backbone with an updating feature-based prototype queue (\textit{i.e.} the gallery queue), and achieve significant improvement on shallow face learning.
We name this training scheme as Semi-Siamese Training, which can be integrated with any existing loss functions and network architectures. As shown in Section~\ref{section_sss_with_various_losses}, whatever loss function, a large improvement can be obtained by using the proposed method for shallow face learning. 

Moreover, we conduct two extra experiments to demonstrate more advantage of SST in a wide range. (1) Although SST is proposed for the shallow data problem, an experiment on conventional deep data shows that leading performance can still be obtained by using SST. (2) Another experiment for verifying the effectiveness of SST for real-world scenario, with pretrain-finetune setting,  also shows that SST outperforms the conventional training.

In summary, the paper includes the following contributions:
\begin{itemize}
\item We formally depict a critical problem of face recognition, \textit{i.e.} Shallow Face Learning, from which the training of face recognition suffers severely. This problem exists in many real-world scenarios but has been overlooked before.
\item We study the Shallow Face Learning problem with thorough experiments, and find the lack of intra-class diversity impedes the optimization and leads to the collapse of the feature space. In such situation, the model suffers from degeneration and over-fitting in the training.
\item We propose Semi-Siamese Training (SST) method to address the issues in Shallow Face Learning. SST is able to perform with flexible combination with the existing loss functions and network architectures.
\item We conduct comprehensive experiments to show the significant improvement by SST on Shallow Face Learning. Besides, the extra experiments show SST also prevails in both conventional deep data and pretrain-finetune task.
\end{itemize}

\section{Related Work}
\subsection{Deep Face Recognition}
There are two major schemes in the deep face recognition. On one hand, the classification based methods is developed from softmax loss and its variants. SphereFace~\cite{liu2017sphereface} introduces the angular margin to enlarge gaps between classes. CosFace~\cite{wang2018cosface} and AM-softmax~\cite{wang2018additive} propose an additive margin to the positive logit. ArcFace~\cite{deng2019arcface} employs an additive angular margin inside the cosine and gives a more clear geometric interpretation. 
On the other hand, the feature embedding methods, such as Contrastive loss~\cite{chopra2005learning,hadsell2006dimensionality,sun2014deep} and Triplet loss~\cite{schroff2015facenet} calculate pair-wise Euclidean distance and optimize the network over the relation between samples pairs or triplets.
N-pairs loss~\cite{sohn2016improved} optimizes positive and negative pairs following a local softmax formulation each mini-batch.
Beyond the two schemes, Zhu \textit{et al.}~\cite{zhu2019large} proposes a classification-verification-classification training strategy and DP-softmax loss to progressively enhance the performance on ID versus spot face recognition task. 

\subsection{Low-shot Face Recognition}
Low-shot Learning (LSL) in face recognition aims at close-set ID recognition by few face samples. Choe \textit{et al.}~\cite{choe2017face} use data augmentation and generation methods to enlarge the training dataset. Cheng \textit{et al.}~\cite{cheng2017know} propose an enforced softmax that contains optimal dropout, selective attenuation, $L_2$ normalization and model-level optimization. Wu \textit{et al.}~\cite{wu2017low} develop the hybrid classifiers by using a CNN and a nearest neighbor model. Guo \textit{et al.}~\cite{guo2017one} propose to align the norms of the weight vectors of the one-shot classes and the normal classes. Yin \textit{et al.}~\cite{yin2019feature} augment feature space of low-shot classes by transferring the principal components from normal to low-shot classes.
The above methods focus on the MS-Celeb-1M Low-shot Learning benchmark~\cite{guo2017one}, which has relatively sufficient samples for each ID in a base set and only one sample for each ID in a novel set, and the target is to recognize faces from both the base and novel set. 
However, as discussed in the previous section, the differences between Shallow Face Learning and LSL have two aspects. First, the LSL methods aim at close-set classification, for example, in the MS-Celeb-1M Low-shot Learning benchmark, the test IDs are included in the training set; but Shallow Face Learning includes open-set recognition where the test samples belong to unseen classes. Second, unlike the LSL generally employing transfer learning from source dataset (pretraining) to target low-shot dataset (finetuning), Shallow Face Learning argues to train from scratch on target shallow dataset. 

\subsection{Self-supervised Learning}
The recent self-supervised methods~\cite{dosovitskiy2014discriminative,wu2018unsupervised,zhuang2019local,he2019momentum} have achieved exciting progress in visual representation learning. Exemplar CNN~\cite{dosovitskiy2014discriminative} introduces the surrogate class concept for the first time, which adopts a parametric paradigm during training and test. Memory Bank~\cite{wu2018unsupervised} formulates the instance-level discrimination as a metric learning problem, where the similarity between instances are calculated from the features in a non-parametric way.  MoCo~\cite{he2019momentum} proposes a dynamic dictionary with a queue and a momentum-updating encoder, which can build a large and consistent dictionary on-the-fly that facilitates the contrastive unsupervised learning. These methods regard each training sample as an instance-level class. Although they employ the data augmentation for each sample, the instance-level classes still lack the intra-class diversity, which is similar to the Shallow Face Learning problem. Inspired by the effectiveness of the self-supervised learning methods, we tackle the issues in Shallow Face Learning with similar techniques, such as the moving-average updating for the Semi-Siamese backbone, and the prototype queue for the supervised loss. 
Nonetheless, SST is quite different with the self-supervised methods. For example, the gallery queue of SST is built based on the gallery samples rather than the sample augmentation technique; SST aims to deal with Shallow Face Learning which is a specific task in supervised learning.
From the perspective of learning against the lack of intra-class diversity, 
our method generalize the advantages of the self-supervised scheme to the supervised scheme on shallow data.

\section{The Proposed Approach}
\subsection{Shallow Face Learning Problem}
\label{problem}
Shallow face learning is a practical problem in real-world face recognition scenario. 
For example, in the authentication application, the face data usually contains a registration photo (gallery) and a spot photo (probe) for each ID.
The ID number could be large, but the shallow depth leads to extreme lack of intra-class information. 
Here, we study how the current classification-based  methods suffer from this problem, and what the consequence is brought by the shallow data.

Most of the current prevailing methods are developed from softmax or its variants, which includes a FC layer, the softmax function, and the cross-entropy loss. The output of the FC layer is the inner product ${w}_{j}^{\mathrm{T}}{x}_{i}$ of the ${i}$-th sample feature ${{x}_{i}}$ and ${j}$-th class weight ${w}_{j}$. When the feature and weight are normalized by their $L_2$ norm, the inner product equals to the cosine similarity ${w}_{j}^{\mathrm{T}}{x}_{i}=\cos{\left(\theta_{{i},{j}}\right)}$. 
Without loss of generality, we take the conventional softmax as an example, and the loss function (omitting the bias term) can be formulated by 
\begin{equation}
\mathcal{L}= -\frac{1}{N} \sum_{i=1}^{N}\log \frac{e^{s \cos \left(\theta_{{i},{y}}\right)}}{e^{s \cos \left(\theta_{{i},{y}}\right)}+\sum_{j=1,j\neq y}^{n} e^{s \cos \left(\theta_{{i},{j}}\right)}},
\label{original_softmax}
\end{equation}
where ${N}$ is the batch size, ${n}$ is the class number, $s$ is the scaling parameter, and $y$ is the ground truth label of the ${i}$-th sample.
The learning objective is maximizing the intra-class pair similarity ${w}_{y}^{\mathrm{T}}{x}_{i}$ and minimizing the inter-class pairs ${w}_{j}^{\mathrm{T}}{x}_{i}$ to achieve compact features for intra-class and separate for inter-class. 
The term inside the logarithm is the prediction probability on the ground truth class $P_y = \frac{e^{s \cos \left(\theta_{{i},{y}}\right)}}{e^{s \cos \left(\theta_{{i},{y}}\right)}+\sum_{j=1,j\neq y}^{n} e^{s \cos \left(\theta_{{i},{j}}\right)}}$, which can be written as $\frac{P_y}{1-P_y} = \frac{e^{s \cos \left(\theta_{{i},{y}}\right)}}{\sum_{j=1,j\neq y}^{n} e^{s \cos \left(\theta_{{i},{j}}\right)}}$.
This equation implies that the optimal solution of the prototype $w_{y}$ satisfies two conditions,
\begin{equation}
\left\{
     \begin{array}{lr}
     w_{y} = \frac{1}{n_y} \sum_{i=1}^{n_y} x_i, & \qquad (i)  \\
     w_j^{\mathrm{T}}{x}_{i} \vert _{j \neq y}=0, & \qquad  (ii)
     \end{array}
\right.
\label{optimal_condition}
\end{equation}
where $n_y$ is the sample number in this class.
The Condition (i) means, ideally, the optimal prototype~${{w}_{y}}$ will be the class center which equals to the average of the features in this classes. Meanwhile, the Condition (ii) pushes the prototype ${w}_{y}$ to the risk of collapse to zeros in many dimensions. When $n_y$ is large enough (deep data), $x_i$'s have large diversity, so keeping the prototype $w_{y} = \frac{1}{n_y} \sum_{i=1}^{n_y} x_i$ away from collapse. 
While in shallow data ($n_y = 2$), the prototype~${{w}_{y}}$ is determined by only two samples in a class, \textit{i.e.} the gallery $x_g$ and probe $x_p$. 
As a result, the three vectors $w_y$, $x_y$ and $x_p$ will rapidly become very close ($w_y \approx x_g \approx x_p$), and this class will achieve very small loss value.
Considering the network is trained batch-wisely by SGD, in every iteration the network is well-fitted on a small number of classes and badly-fitted on the other classes, thus the total loss value will be oscillating and the training will be harmed (as shown in Fig.~\ref{loss_oscillation} dot curves).
Moreover, since all the classes gradually lose the intra-class diversity in features space $x_g \approx x_p$, the prototype ${{w}_{y}}$ is pushed to zeros in most dimensions by Condition (ii), and unable to span a discriminative feature space.

\setlength{\tabcolsep}{2pt}
\begin{table}[t]
\begin{center}
\caption{The performance(\%) on training data and LFW test.}
\label{problem_formulation}
\begin{tabular}{|c|l|l|l|l|l|l|l|l|}
\hline
\multirow{2}{*}{Data}&
\multicolumn{2}{c|}{Softmax}&
\multicolumn{2}{c|}{A-softmax}&
\multicolumn{2}{c|}{AM-softmax}&
\multicolumn{2}{c|}{Arc-softmax}
\\ \cline{2-9}&Training&Test&Training&Test&Training&Test&Training&Test\\
\hline\hline
Deep&99.83&99.10&99.96&99.38&99.42&99.32&98.74&99.40\\
\hline
Shallow&99.40~\textcolor{red}{$\downarrow$}&92.64~\textcolor{red}{$\downarrow$}&99.42~\textcolor{red}{$\downarrow$}&94.67~\textcolor{red}{$\downarrow$}&99.98~\textcolor{green}{$\uparrow$}&92.75~\textcolor{red}{$\downarrow$}&99.99~\textcolor{green}{$\uparrow$}&94.32~\textcolor{red}{$\downarrow$}\\
\hline
\end{tabular}
\end{center}
\end{table}

\begin{figure}[t]
\centering
\begin{minipage}[t]{0.375\linewidth}
\includegraphics[height=3cm]{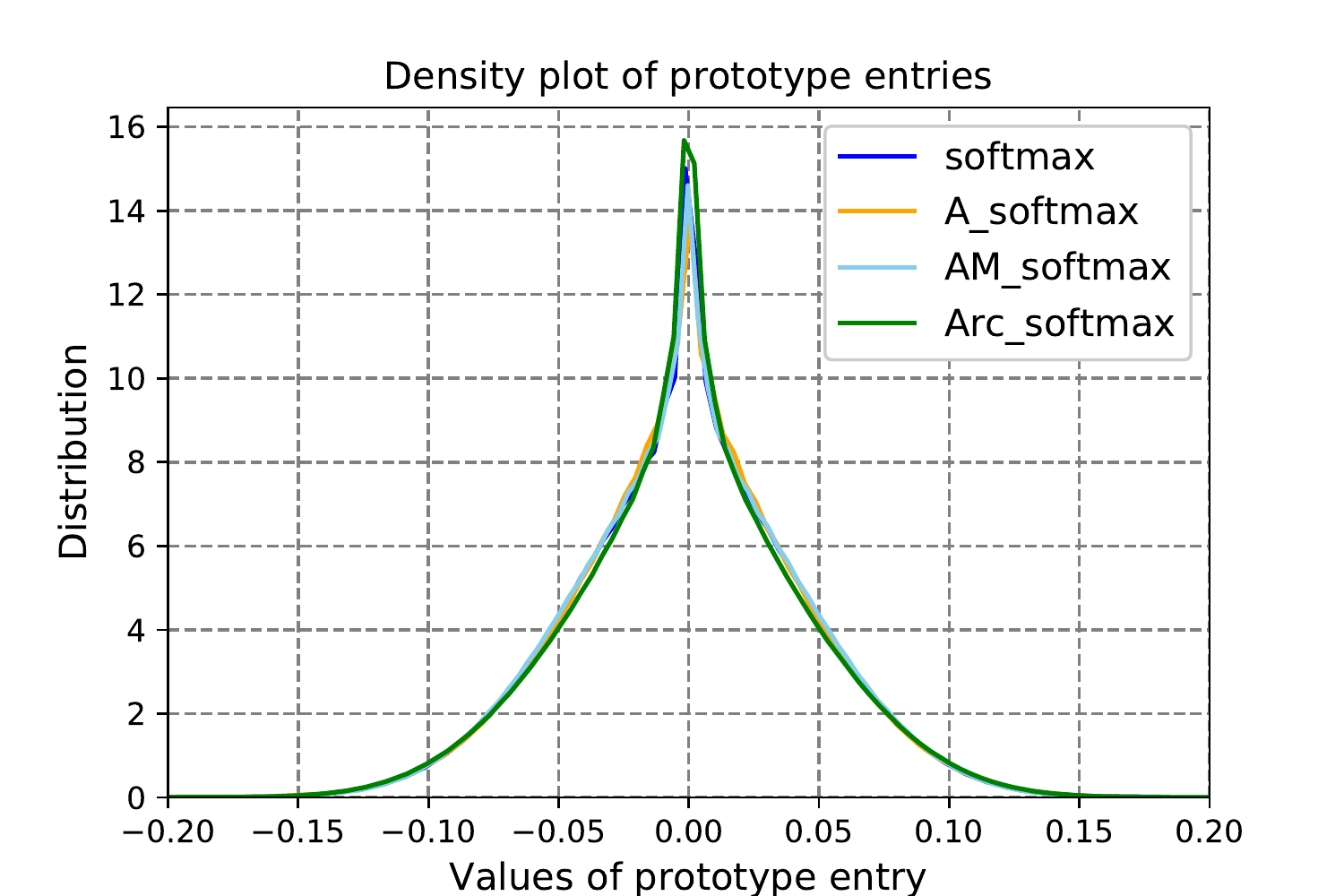}
\subcaption{}
\label{distribution_a}
\end{minipage}
\begin{minipage}[t]{0.375\linewidth}
\includegraphics[height=3cm]{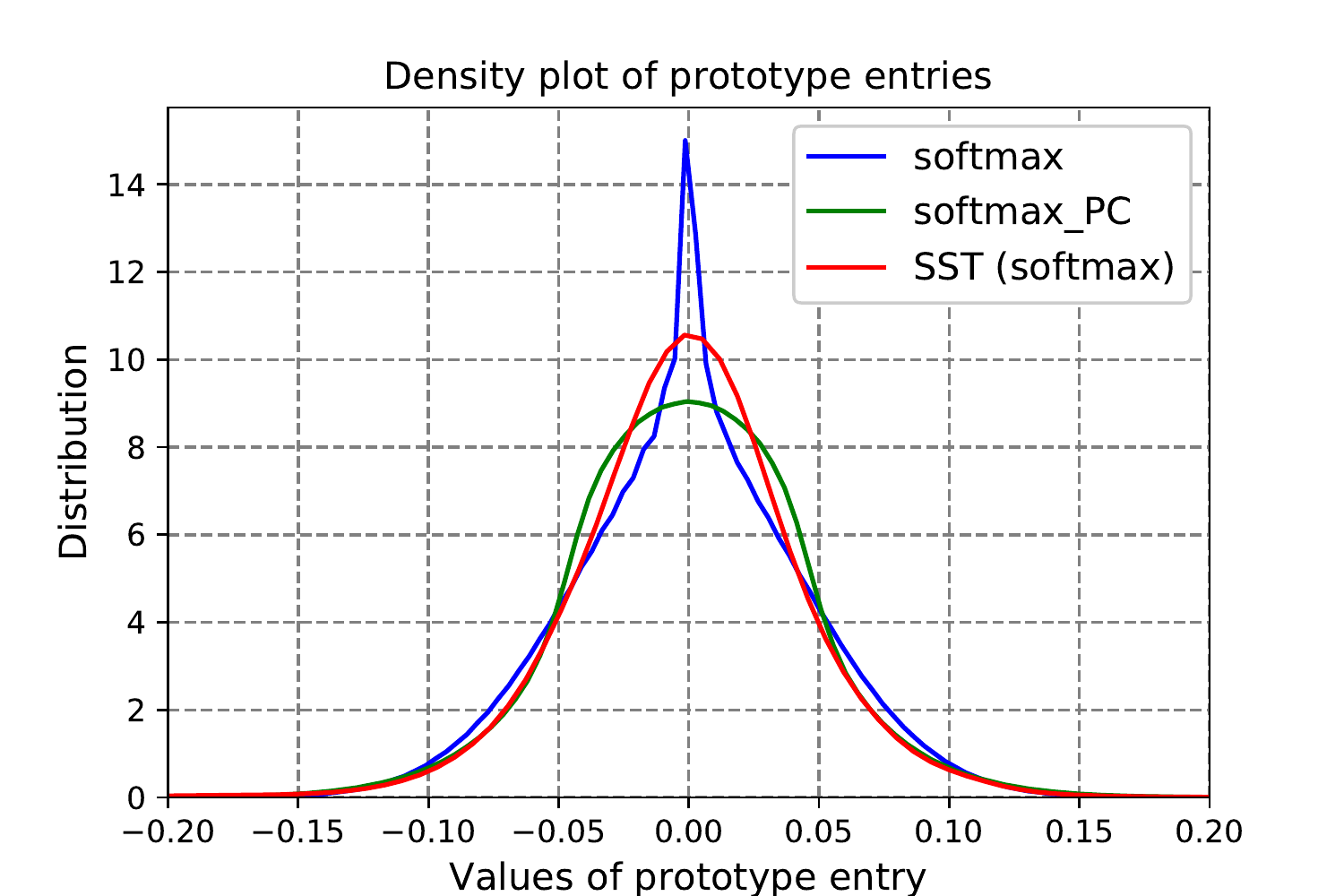}
\subcaption{}
\label{distribution_b}
\end{minipage}

\caption{
(a) The distribution of the entry values of prototype $w_y$ by the conventional training.
The loss functions are softmax, A-softmax, AM-softmax and Arc-softmax.
(b) The red curve is the distribution of the entry values of prototype $x_y$ by SST. The green curve is that by prototype constraint. The loss function is softmax. Best viewed in color.}
\end{figure}

To explore the consequence brought by the shallow data problem, we conduct an experiment on both deep data and shallow data with the loss functions of softmax,  A-softmax~\cite{liu2017sphereface}, AM-softmax~\cite{wang2018additive} and Arc-softmax~\cite{deng2019arcface}.
The deep data is MS1M-v1c~\cite{trillionpairs.org} (cleaned version of MS-Celeb-1M~\cite{guo2016ms}). Shallow data is a subset of MS1M-v1c, with two face images selected randomly per ID from the deep data.
Table~\ref{problem_formulation} shows not only the test accuracy on LFW~\cite{huang2008labeled} but also the accuracy on the training data.
We can find that the softmax and A-softmax get lower performance both in training and test when training data becomes from deep to shallow, 
while the AM-softmax and Arc-softmax get higher in training but lower in test.
Therefore, we argue that the softmax and A-softmax suffer from the model degeneration issue, while the AM-softmax and Arc-softmax suffer from the over-fitting issue.
To further support this argument, we inspect the value of each entry in the prototype ${{w}_{y}}$, and compute the distribution with Parzen window. The distribution is displayed in Fig~\ref{distribution_a}, with the horizontal axis represents the entry values, and the vertical axis represents the density. 
We can find that most entries of the prototypes degrade to zeros, which means the feature space collapses in most dimensions. In such reduced-dimension space, the models could be easily degenerated or over-fitted.

\subsection{Semi-Siamese Training}

\begin{figure}[t]
\centering
\includegraphics[scale=0.25]{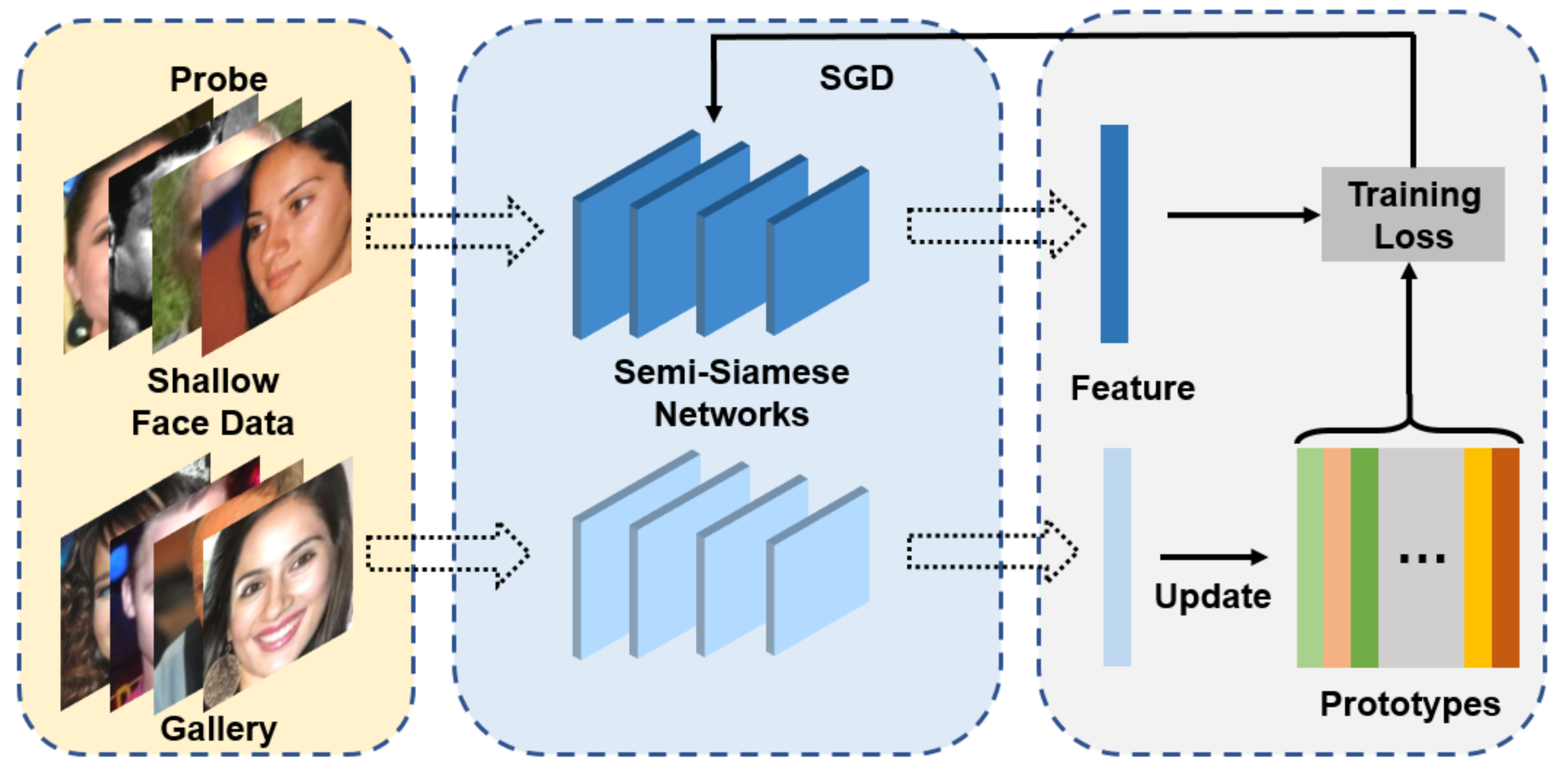}
\caption{The overview of Semi-Siamese Training (SST). SST includes a pair of Semi-Siamese networks, which have a probe-set network (the top dark blue network) to embed the probe features, and a gallery-set network (the bottom pale blue network) to update prototypes by gallery features. SST employs the probe features and the feature-based prototypes to compute the training losses which can be any existing loss such as softmax, Arc-softmax, Contrastive \textit{etc}. 
Finally, the probe-set network is optimized via SGD \textit{w.r.t.} the training loss, and the gallery-set network is updated by the moving-average. Best viewed in color.}
\label{definiton}
\end{figure}

From the above analysis, we can see, when the data becomes shallow, the current methods are damaged by the model degeneration and over-fitting issues, and the essential reason consists in feature space collapse. To cope with this problem, there are two directions for us to proceed: (1) to make $w_y$ and $x_i$ updating correctly, and (2) to keep the entries of $w_y$ away from zeros.

In the first direction, the major issue is the network is prevented from effective optimization. We retrospect the Condition (i) in Eqn.~\ref{optimal_condition} for Shallow Face Learning in which only two face images are available for each ID. We denote them by $I_g$ (gallery) and $I_p$ (probe) and their features $x_g = \phi (I_g)$ and $x_p = \phi (I_p)$, where $\phi$ is the Siamese backbone. According to Condition (i), $w_{y} = \frac{1}{2} (x_g + x_p)$. Due to the lack of intra-class diversity, the gallery and probe often have close features, and thus $w_y = \frac{1}{2} (x_g + x_p) \approx x_g \approx x_p$. As studied in the previous subsection, this situation will lead to loss value oscillation, preventing the network from effective optimization. The basic idea to deal with the problem is to keep $x_g$ some distance from $x_p$, \textit{i.e.} $x_g = x_p + \epsilon, \forall \epsilon > 0$. To maintain the distance between $x_g$ and $x_p$, we propose to make the Siamese backbone $\phi$ being Semi-Siamese. Specifically, a gallery-set network $\phi_g$ gets input of gallery, and a probe-set network $\phi_p$ gets input of probe. $\phi_g$ and $\phi_p$ have the same architecture but non-identical parameters, $\phi_g = \phi_p + \epsilon'$, so the features prevent being attracted to each other $\phi_g(I_g) = \phi_p(I_p) + \epsilon$. There are certain choices to implement the Semi-Siamese networks. 
For example, one can add a network constraint $\| \phi_g - \phi_p \| < \epsilon'$ in the training loss, such as $\mathcal{L}_{\text{total}}=\mathcal{L} + \lambda *\left\|\phi_{g}-\phi_{p}\right\|$, and the non-negative parameter $\lambda$ is used to balance the network constraint in the training loss. Another choice, as suggested by MoCo~\cite{he2019momentum}, aims to update the gallery-set network in the momentum way,
\begin{equation}
{\phi}_{g} = {m}\cdot{\phi}_{g}+{(1-m)}\cdot{\phi}_{p},
\label{moving_average}
\end{equation}
where $m$ is the weight of moving-average, and the probe-set network $\phi_p$ updates with SGD \textit{w.r.t.} the training loss. Both $ \lambda$ and $m$ are the instantiation of $\epsilon'$ which keeps $\phi_g$ and $\phi_p$ similar.
We compare different implementations for the Semi-Siamese networks, and find the moving-average style gives significant improvement in the experiments. Owing to the intra-class diversity maintaining, the training loss decreases steadily without oscillation (solid curves in Fig.~\ref{loss_oscillation}).

In the second direction, a straightforward idea is to add a prototype constraint in the training loss to enlarge the entries of prototype, such like $\mathcal{L} + \beta (\alpha - \|w_y\|)$ with parameters $\alpha$ and $\beta$. However, we find this technique enlarges the entries in most dimension indiscriminately (Fig.~\ref{distribution_b} the green distribution), and results in decrease (Table~\ref{ablation}). Instead of manipulating $w_y$, we argue to replace $w_y$ by the gallery feature $x_g$ as the prototype. Thus, the prototype totally depends on the output of the backbone, avoiding the zero issue of the parameters (entries) of $w_y$.
The red distribution in Fig.~\ref{distribution_b} shows the feature-based prototype avoids the issue of collapse while keeping more discriminative components compared with the prototype constraint.
Removing $w_y$ also alleviates the over-fitting risk of heavy parameters. 
The entire prototype set updates by maintaining a gallery queue.
Certain self-learning methods~\cite{wu2018unsupervised,he2019momentum} have studied this technique and its further advantages, such as better generalization when encountering unseen test IDs. 

In summary, our Semi-Siamese Training method is developed to address the Shallow Face Learning problem along the two directions.
The forward propagation backbone is constituted by a pair of Semi-Siamese networks, each of which is in charge of feature encoding for gallery and probe, respectively; 
the training loss is computed with an updating gallery queue, so the networks are optimized effectively on the shallow data. 
This training scheme can be integrated with any form of existing loss function (no matter classification loss or embedding loss) and network architectures (Fig.~\ref{definiton}).

\section{Experiments}
This section is structured as follows. Section~\ref{section_exp_datasets} introduces the datasets and experimental settings. Section~\ref{section_ablation_study} includes the ablation study on SST. Section~\ref{section_sss_with_various_losses} demonstrates the significant improvement by SST on Shallow Face Learning with various loss functions. Section~\ref{section_sss_with_various_nets} shows the convergence of SST with various backbones. Section~\ref{section_sss_on_deep_data} shows SST can also achieve leading performance on deep face data. Section~\ref{Pretrain_and_Finetune} studies SST also outperforms conventional training for the pretrain-finetune task.
\subsection{Datasets and Experimental Settings}
\label{section_exp_datasets}


\textbf{Training Data.}
To prove the reproducibility\footnote[1]{The source code of SST is available at https://github.com/dituu/Semi-Siamese-Training.}, we employ the public datasets for training. 
To construct shallow data, two images are randomly selected for each ID from the MS1M-v1c~\cite{trillionpairs.org} dataset. Thus, the shallow data includes 72,778 IDs and 145,556 images.
For deep data, we use the full MS1M-v1c which has 44 images per ID in average.
Besides, we utilize a real-world surveillance face recognition benchmark QMUL-SurvFace~\cite{cheng2018surveillance} for the experiment of pretrain-finetune.

\textbf{Test Data.}
For a thorough evaluation, we adopt LFW~\cite{huang2008labeled}, BLUFR~\cite{liao2014benchmark}, AgeDB-30~\cite{moschoglou2017agedb}, CFP-FP~\cite{sengupta2016frontal}, CALFW~\cite{zheng2017cross}, CPLFW~\cite{zheng2018cross}, MegaFace~\cite{kemelmacher2016megaface} and QMUL-SurvFace~\cite{cheng2018surveillance} datasets. AgeDB-30 and CALFW focus on large age gap face verification. CFP-FP and CPLFW aim at cross-pose variants face verification. BLUFR is dedicated for the evaluation with focus at low false accept rates (FAR), and we report the verification rate at the lowest FAR (1e-5) on BLUFR. MegaFace also evaluates the performance of large-scale face recognition with the millions of distractors. QMUL-SurvFace test set aims at real-world surveillance face recognition and has a large domain gap compared to above benchmarks. 
\par
\textbf{Prepossessing.} All face images are detected by the FaceBoxes~\cite{zhang2017faceboxes}. Then, we 
align and crop faces to 144$\times$144 RGB images by five facial landmarks~\cite{feng2018wing}. 
\par
\textbf{CNN Architecture.} To balance the performance and the time cost, we use the MobileFaceNet~\cite{chen2018mobilefacenets} in the ablation study and the experiments with various loss functions. Besides, we employ Attention-56~\cite{wang2017residual} in the deep data and pretrain-finetune experiments. The output is a 512-dimension feature. 
In addition, we also employ extra backbones including VGG-16~\cite{Simonyan2014Very}, SE-ResNet-18~\cite{hu2018squeeze}, ResNet-50 and -101~\cite{he2016deep} to prove the convergence of SST with various architectures.
\par
\textbf{Training and Evaluation.} 
Four NVIDIA Tesla P40 GPUs are employed for training.  The batch size is 256 and the learning rate begins with 0.05. 
In the shallow data experiments, the learning rate is divided by 10 at the 36k, 54k iterations and the training process is finished at 64k iterations.
For the deep data, we divide the learning rate at the 72k, 96k, 112k iterations and finish at 120k iterations. 
For pretrain-finetune experiments, the learning rate starts from 0.001 and is divided by 10 at the 6k, 9k iterations and finished at 10k iterations. 
The size of the gallery queue depends on the number of classes in training datasets, so we empirically set it as 16,384 for shallow and deep data, and 2,560 for QMUL-SurvFace. 
In the evaluation stage, we extract the last layer output from the probe-set network as the face representation. 
The cosine similarity is utilized as the similarity metric. For strict and precise evaluation, all the overlapping IDs between training and test datasets are removed according to the list~\cite{wang2019co}. 
\par
\textbf{Loss Function.} 
SST can be flexible integrated with the existing training loss functions. Both
classification and embedding learning loss functions are considered as the baseline, and compared with the integration with SST. The classification loss functions include A-softmax~\cite{liu2017sphereface}, AM-softmax~\cite{wang2018additive}, Arc-softmax~\cite{deng2019arcface}, AdaCos~\cite{zhang2019adacos}, MV-softmax~\cite{wang2019mis},  DP-softmax~\cite{zhu2019large} and Center loss~\cite{wen2016discriminative}. The embedding learning methods include Contrastive~\cite{sun2014deep}, Triplet~\cite{schroff2015facenet} and N-pairs~\cite{sohn2016improved}. 

\begin{table}[t!]
\caption{Ablation study. Performance ($\%$) on LFW, AgeDB-30, CFP-FP, CALFW, CPLFW and BLUFR. 
}
\label{ablation}

 \begin{minipage}[t]{0.5\textwidth}
  \centering
  \resizebox{\textwidth}{!}{
    \begin{tabular}{|c||c|c|c|c|c|c|}
    \hline
     & LFW & AgeDB & CFP & CALFW & CPLFW & BLUFR\\
    \hline\hline
    \multicolumn{7}{|c|}{softmax}\\
    \hline
         Org.& 92.64 & 73.96 & 70.80 & 73.05 & 62.64 & 27.05\\
         \hline
         A & 91.36 & 71.85 & 69.00 & 72.14 & 61.35 & 24.87\\
         \hline
         B & 93.43 & 76.00 & 71.46 & 74.65 & 62.68 & 30.65\\
         \hline
         C & 96.62 & 82.63 & 79.10 & 80.18 & 67.55 & 52.05\\
         \hline
         D & 98.32 & 88.77 & 84.81 & 86.63 & 74.80 & 69.93\\
         \hline
         SST & \textbf{98.77} & \textbf{91.60} & \textbf{88.63} & \textbf{89.82} & \textbf{78.43} & \textbf{77.58}\\
    \hline\hline
    \multicolumn{7}{|c|}{A-softmax}\\
    \hline
         Org.&94.67 &77.88 & 72.90 & 75.85 & 64.00 & 37.16\\
         \hline
         A & 93.76 & 76.79 & 71.35 & 74.56 & 62.80 & 35.18\\
         \hline
         B & 94.62 & 78.08 & 74.03 & 76.35 & 63.87 & 38.35\\
         \hline
         C & 96.32 & 82.28 & 81.30 & 81.05 & 68.77 & 57.13\\
         \hline
         D & 97.52 & 85.83 & 81.87 & 83.88 & 71.03 & 60.79\\
         \hline
         SST & \textbf{98.98} &\textbf{91.88} & \textbf{89.54} & \textbf{89.73} & \textbf{77.68} & \textbf{80.65}\\
         \hline
    \end{tabular}
    }
  \end{minipage}
  \begin{minipage}[t]{0.5\textwidth}
   \centering
   \resizebox{\textwidth}{!}{
    \begin{tabular}{|c||c|c|c|c|c|c|}
    \hline
     & LFW & AgeDB & CFP & CALFW & CPLFW & BLUFR\\
    \hline\hline
    \multicolumn{7}{|c|}{AM-softmax}\\
    \hline
        Org.& 92.75 & 75.30 & 68.74 & 76.63 & 63.63 & 33.23\\
        \hline
        A & 92.35 & 74.12 & 68.08 & 74.89 & 62.76 & 32.12\\
         \hline
        B & 93.25 & 76.16 & 69.17 & 77.78 & 63.88 & 36.59\\
         \hline
        C & 98.02 & 86.37 & 85.17 & 85.72 & 72.83 & 62.07\\
         \hline
        D & 98.30 & 88.18 & 87.31 & 87.93 & 76.27 & 75.46\\
        \hline
        SST & \textbf{98.97} & \textbf{92.25} & \textbf{88.97}  
        & \textbf{90.23} & \textbf{79.45} &  \textbf{84.95}\\
     \hline\hline
    \multicolumn{7}{|c|}{Arc-softmax}\\
    \hline
        Org.& 94.32 & 77.80 & 71.25 & 78.15 & 65.45 & 40.34\\
        \hline
        A & 93.60 & 77.35 & 70.59 & 77.78 & 64.28& 40.08\\
         \hline
        B & 94.48 & 78.42 & 72.15 & 78.65 & 65.78 & 42.50\\
         \hline
        C & 98.20 & 85.28 & 81.50 & 83.50 & 71.32 & 60.67\\
        \hline
        D & 98.08 & 88.68 & 84.54 & 86.92 & 74.40 & 68.84\\
        \hline
        SST & \textbf{98.95} & \textbf{91.73} & \textbf{88.59}  & \textbf{89.85} &\textbf{79.60} & \textbf{82.74}\\
        \hline
    \end{tabular}
    }
   \end{minipage}
\end{table}

\subsection{Ablation Study}
\label{section_ablation_study}
We analyze each technique in SST, and compare them with the other choices mentioned in the previous section, such as the network constraint ($\| \phi_g - \phi_p \| <  \epsilon'$) and the prototype constraint ($\beta (\alpha - \|w_y\|)$).
Table~\ref{ablation} compares their performance with four basic loss functions (softmax, A-Softmax, AM-Softmax and Arc-softmax).
In this table, ``Org." denotes the plain training, 
``A" denotes the prototype constraint, 
``B" denotes the network constraint, 
``C" denotes the gallery queue, 
``D" denotes the combination of ``B" and ``C", 
``SST" denotes the ultimate scheme of Semi-Siamese Training which includes the moving-average updating Semi-Siamese networks and the training scheme with gallery queue. 
From Table~\ref{ablation}, we can conclude: (1) the naive prototype constraint ``A" leads to decrease in most terms, which means enlarging $w_y$ in every dimension indiscriminatively does not help on Shallow Face Learning; (2) the network constraint ``B" and the gallery queue ``C" results in progressive increase, and the combination of them ``D" obtains further improvement; (3) finally, SST employs moving-average updating and gallery queue, and achieves the best results by all terms. The comparison indicates SST well addresses the problem in Shallow Face Learning, and obtains significantly improvements in test accuracy.

\par
\subsection{SST with Various Loss Functions}
\label{section_sss_with_various_losses}

\begin{figure}[t!]
\centering
\includegraphics[height=3.8cm]{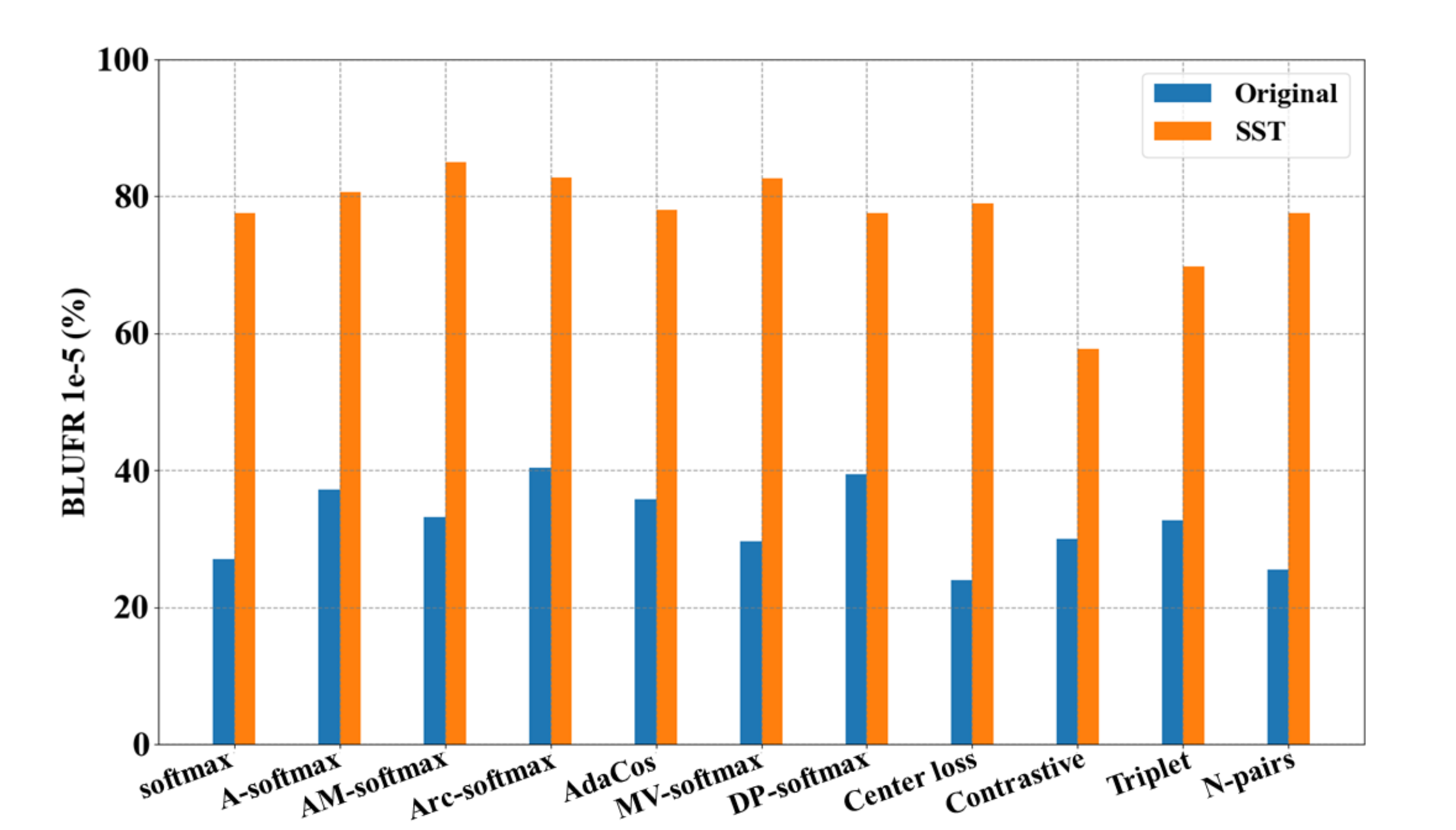}
\caption{After being integrated with SST, every loss function obtains large increase on Shallow Face Learning.
The blue bars correspond the conventional training on shallow data, the orange bars correspond to SST on shallow data. 
The test results are the verification rates at FAR=1e-5 on BLUFR.
Best viewed in color.}
\label{various_loss}
\end{figure}

First, we train the network on the shallow data with various loss functions and test it on BLUFR at FAR=1e-5 (the blue bars in Fig.~\ref{various_loss}).
The loss functions include classification and embedding ones such as softmax, A-softmax, AM-softmax, Arc-softmax, AdaCos, MV-softmax, DP-softmax, Center loss, Contrastive, Triplet and N-pairs. 
Then, we train the same network with the same loss functions on the shallow data, but with SST scheme. As shown in Fig.~\ref{various_loss}, SST can be flexibly integrated with every loss function, and obtains large increase for Shallow Face Learning (the orange bars).
Moreover, we employ hard example mining strategies when training on MV-softmax and embedding losses. The results prove SST can also work well with the hard example mining strategies.

\subsection{SST with Various Network Architectures}
\label{section_sss_with_various_nets}

\begin{figure}[t]
\centering
\begin{minipage}[t]{0.475\linewidth}
\includegraphics[height=2.9cm]{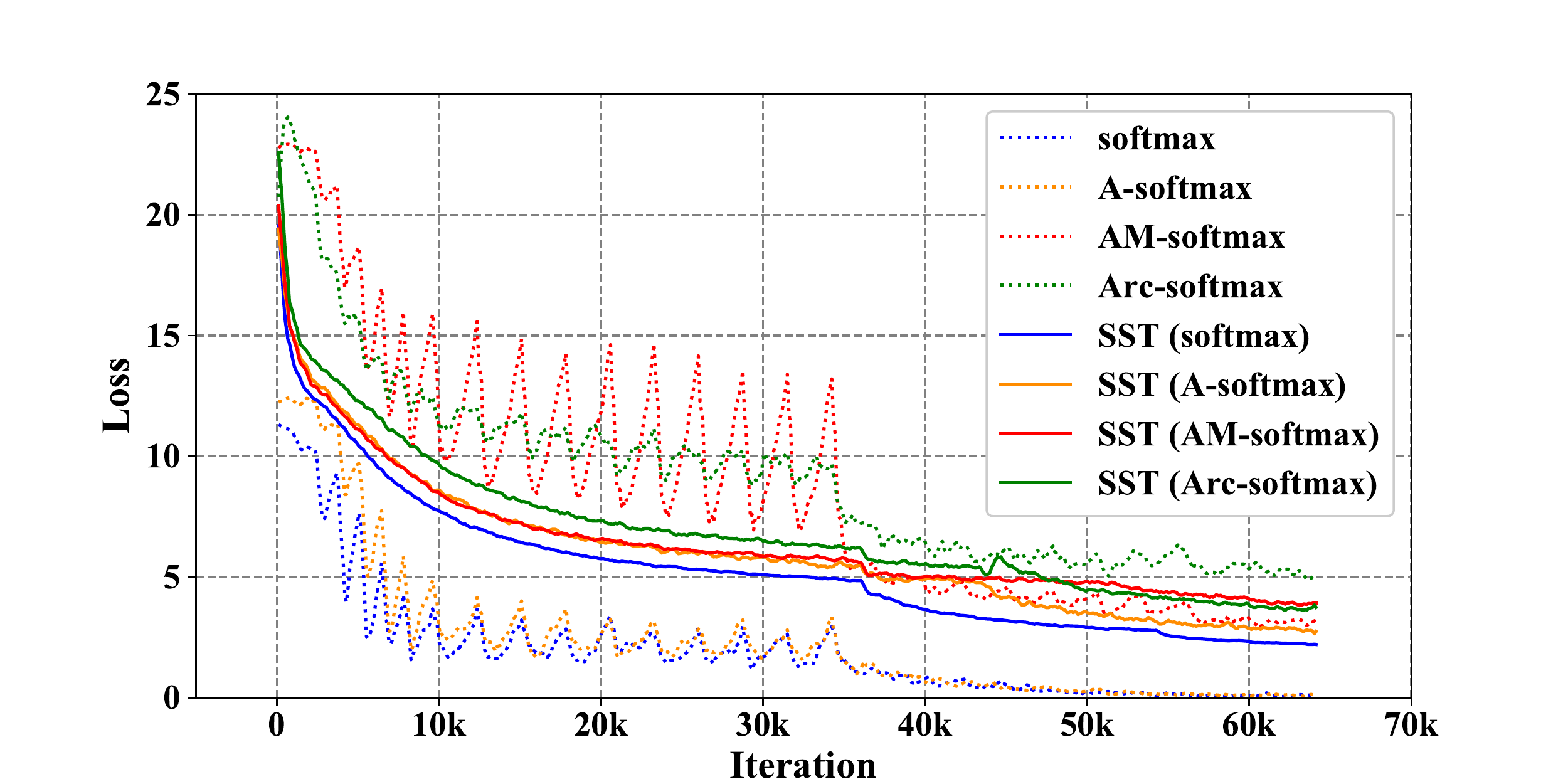}
\subcaption{}
\label{loss_oscillation}
\end{minipage}
\begin{minipage}[t]{0.475\linewidth}
\includegraphics[height=2.9cm]{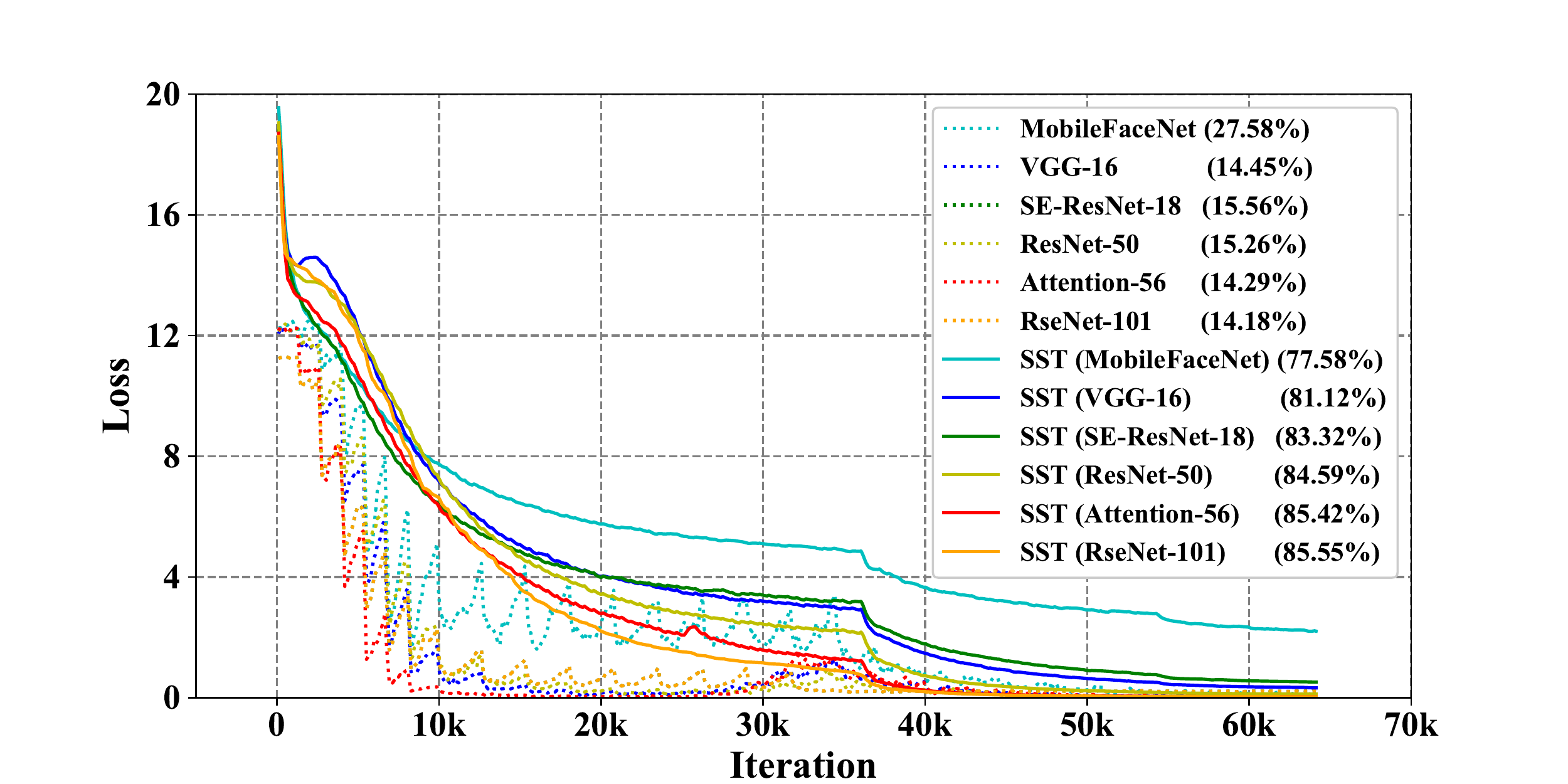}
\subcaption{}
\label{loss_curves}
\end{minipage}
\caption{
(a) The loss values of conventional training (the dot curves) and SST (the solid curves) with different training loss functions along training iteration. By maintaining the intra-class diversity, SST can prevent the oscillation and achieve steady convergence.
(b) The loss values of conventional training and SST with various network architectures along training iteration. The digits in the legend are the test accuracy of each network on BLUFR.
Best viewed in color.}
\end{figure}

To demonstrate the stable convergence in the training, we employ SST to train different CNN architectures, including
MobileFaceNet, VGG-16, SE-ResNet-18, Attention-56, ResNet-50 and -101. As shown in Fig.~\ref{loss_curves}, the loss curves of conventional training (the dot curves) suffer from oscillation. But every loss curve of SST (the solid curves) decreases steadily, indicating the convergence of each network along with the training of SST.
Besides, the digits in the legend of Fig.~\ref{loss_curves} indicates the test result of each network on BLUFR. 
For conventional training, the test accuracy decreases with the deeper network architectures, showing that the larger model size exacerbates the model degeneration and over-fitting. 
In contrast, as the network becomes heavy, the test accuracy of SST increases, showing that SST makes increasing contribution with more complicated architectures.

\subsection{SST on Deep Data Learning}
\label{section_sss_on_deep_data}

\setlength{\tabcolsep}{1pt}
\begin{table}[t!]
\begin{center}
\caption{
Comparison of Semi-Siamese Training and the conventional training on deep data. In MegaFace, ``Id.'' refers to face identification rank1 accuracy with 1M distractors, and ``Veri.'' refers to face verification rate at 1e-6 FAR.}
\begin{tabular}{|c|p{0.9cm}<{\centering}|p{1.1cm}<{\centering}|p{0.9cm}<{\centering}|p{1.25cm}<{\centering}|p{1.2cm}<{\centering}|p{1.2cm}<{\centering}|p{1cm}<{\centering}|p{1cm}<{\centering}|}
\hline
\multirow{2}{*}{Method}&
\multirow{2}{*}{LFW}&
\multirow{2}{*}{AgeDB}&
\multirow{2}{*}{CFP}&
\multirow{2}{*}{CALFW}&
\multirow{2}{*}{CPLFW}&
\multirow{2}{*}{BLUFR}&
\multicolumn{2}{c|}{MegaFace}
\\ \cline{8-9}&&&&&&& Id. &Veri.\\
\hline\hline
softmax&99.58&95.33&92.66&93.18&84.47&93.15&89.89&92.00\\
AM-softmax&99.70&97.03&94.17&94.41&87.00&94.25&95.67&96.35\\
Arc-softmax&99.73&97.18&94.37& \textbf{95.25}&87.05& \textbf{95.29}&96.10&96.81\\
\hline
SST(softmax)&99.67&96.37&94.96&94.18&85.82&94.56&91.01&93.23\\
SST(AM-softmax)&99.75& \textbf{97.20}&95.10&94.62& \textbf{88.35}&94.84& \textbf{96.27}& \textbf{96.96}\\
SST(Arc-softmax)& \textbf{99.77}&97.12& \textbf{95.96}&94.78&87.15&94.76&95.63&96.50\\

\hline
\end{tabular}
\label{results_on_deep_data}
\end{center}
\end{table}

The previous experiments show SST has well tackled the problems in Shallow Face Learning and obtained significant improvement in test accuracy. To further explore the advantage of SST for wider application, we adopt SST scheme on the deep data (full version of MS1M-v1c), and make comparison with the conventional training. 
Table~\ref{results_on_deep_data} shows the performance on LFW, AgeDB-30, CFP-FP, CALFW, CPLFW and BLUFR and MegaFace.
SST gains the leading accuracy in most of the test sets, and also the competitive results on CALFW and BLUFR.
SST (softmax) achieves at least one percent improvement on AgeDB-30, CFP-FP, CALFW and CPLFW which include the hard cases of large face pose or large age gap.
Notably, SST reduces large amount of FC parameters by which the classification loss is computed for the conventional training. One can refer to the supplementary material for more results on deep data.

\begin{table}[t!]
\begin{center}
\caption{QMUL-SurvFace evaluation. ``TPR(\%)@FAR" includes the true positive verification rate at varying FARs, and ``TPIR20(\%)@FPIR" includes rank-20 true positive identification rate at varying false positive identification rates. 
}
\label{Finetune}
\begin{tabular}{|c|p{1cm}<{\centering}|p{1cm}<{\centering}|p{1cm}<{\centering}|p{1cm}<{\centering}|p{1cm}<{\centering}|p{1cm}<{\centering}|p{1cm}<{\centering}|p{1cm}<{\centering}|}
\hline
\multirow{2}{*}{Method}&
\multicolumn{4}{c|}{TPR(\%)@FAR}&
\multicolumn{4}{c|}{TPIR20(\%)@FPIR}
\\ \cline{2-9}&0.3&0.1&0.01&0.001&0.3&0.2&0.1&0.01\\
\hline\hline
softmax &73.09&52.29&26.07&12.54&8.09&6.25&3.98&1.13\\
AM-softmax~\cite{wang2018additive} &69.59&47.67&23.90&13.24&9.07&7.14&4.65&1.34\\
Arc-softmax~\cite{deng2019arcface} &68.14&48.65&24.12&11.34&8.77&6.88&4.79&1.36\\
DP-softmax~\cite{zhu2019large} &76.32&55.85&25.32&11.64&7.50&5.38&3.38&0.95\\
Contrastive~\cite{sun2014deep} &84.48&67.99&31.87&5.31&9.16&6.91&4.44&0.10\\
Triplet~\cite{schroff2015facenet}
&85.59&69.61&33.76&7.20&10.14&7.70&4.75&0.37\\
N-pairs~\cite{sohn2016improved}
&87.26&67.04&29.67&12.07&10.75&8.09&4.87&0.41\\
\hline
SST(softmax) &81.08&63.41&34.20&19.03&11.24&8.49&5.28&1.21\\
SST(AM-softmax)  &86.49&69.41&36.21&18.51&12.22&9.51&5.85&1.63\\
SST(Arc-softmax) &87.00&68.21&35.72&\textbf{22.18}&\textbf{12.38}&\textbf{9.71}&\textbf{6.61}&\textbf{1.72}\\
SST(DP-softmax) &87.69&69.69&\textbf{36.32}&14.83&10.20&7.83&5.14&1.08\\
SST(Contrastive) &87.54&69.91&32.15&9.58&9.87&7.38&4.76&0.78\\
SST(Triplet)  &\textbf{90.65}&\textbf{73.35}&33.85&12.48&11.09&8.14&5.27&0.92\\
SST(N-pairs) &89.31&71.26&32.34&15.96&11.30&9.13&5.68&1.22\\
\hline
\end{tabular}
\end{center}
\end{table}


\subsection{Pretrain and Finetune}
\label{Pretrain_and_Finetune}
In real-world face recognition, there is a large domain gap between the public training datasets and the captured face images.
The public training datasets, such as MS-Celeb-1M and VGGFace2, are well-posed face images collected from internet. But the real-world applications are usually quite different. 
To cope with this issue, the typical routine is to pretrain a network on the public training datasets and fine-tune it on real-world face data. 
Although SST is dedicated to the training from scratch on shallow data, we are still interested in employing SST to deal with the challenge in finetuning task. So, we conduct an extra experiment with pretraining on MS1M-v1c and finetuning on QMUL-SurvFace in this subsection. The network is first pretrained with softmax on MS1M-v1c. We randomly select two samples for each ID from the QMUL-SurvFace to construct the shallow data. The network is then finetuned on QMUL-SurvFace shallow data with/without SST. The evaluation is performed on the QMUL-SurvFace test set. 
From Table~\ref{Finetune}, we can find that, no matter for classification learning or embedding learning, SST boosts the performance significantly in both verification and identification, compared with the conventional training.


\section{Conclusions}
In this paper, we first study a critical problem in real-world face recognition, \textit{i.e.} Shallow Face Learning, which has been overlooked before. We analyze how the existing training methods suffer from Shallow Face Learning.
The core issues consist in the training difficulty and feature space collapse, which leads to the model degeneration and over-fitting.
Then, we propose a novel training method, namely Semi-Siamese Training (SST), to address challenges in Shallow Face Learning. Specifically, SST employs the Semi-Siamese networks and constructs the gallery queue with gallery features to overcome the issues. SST can perform with flexible integration with the existing training loss functions and network architectures. Experiments on shallow data show SST significantly improves the conventional training. Besides, extra experiments further explore the advantage of SST in a wide range, such as deep data training and pretrain-finetune development.

\section*{Acknowledgement}
This work was supported in part by the National Key Research \& Development Program (No. 2020YFC2003901), Chinese National Natural Science Foundation Projects \#61872367, and \#61572307,
and Beijing Academy of Artificial Intelligence (BAAI).

\clearpage

\title{Semi-Siamese Training for Shallow Face Learning \\ Supplementary Material} 

\titlerunning{Semi-Siamese Training for Shallow Face Learning}
\authorrunning{H. Du, et al.}
\author{Hang Du\inst{1,2}\thanks{Equal contribution. This work was performed at JD AI Research.} 
\and Hailin Shi\inst{2}\printfnsymbol{1} 
\and Yuchi Liu\inst{2}
\and Jun Wang\inst{2} 
\and Zhen Lei\inst{3}
\and \\ Dan Zeng\inst{1}\textsuperscript{\Letter}
\and Tao Mei\inst{2} }
\institute{
Shanghai University, Shanghai, China. \\
\email{\{duhang, dzeng\}@shu.edu.cn} \\
\and
JD AI Research, Beijing, China.\\
\email{\{shihailin, wangjun492, tmei\}@jd.com, u6009551@anu.edu.au}\\ 
\and 
Institute of Automation, Chinese Academy of Science, Beijing, China \\
\email{zlei@nlpr.ia.ac.cn}
}
\maketitle


\section{Additional Experiments and Analysis}

\subsection{SST on Deep Data Learning}

First, we provide more details about utilizing SST on deep data learning. In each iteration of deep data training, a batch of ID is randomly sampled, and for each ID, two images are randomly sampled. The arbitrary one acts as gallery, and the other one acts as probe. So, every image has $50\%$ chance to play the role of gallery or probe. Besides, 
as a supplementary experiment for Section 4.5 of the main paper, we evaluate SST with loss functions of DP-softmax~\cite{zhu2019large}, Contrastive~\cite{sun2014deep}, Triplet~\cite{schroff2015facenet} and N-pairs~\cite{sohn2016improved}. For evaluation, we use seven test benchmarks, including LFW~\cite{huang2008labeled}, BLUFR~\cite{liao2014benchmark}, AgeDB~\cite{moschoglou2017agedb}, CFP~\cite{sengupta2016frontal}, CALFW~\cite{zheng2017cross}, CPLFW~\cite{zheng2018cross}, MegaFace~\cite{kemelmacher2016megaface}.
From the results, we can find all the loss functions can achieve better performance on various benchmarks after employing SST. Moreover, we can observe the original embedding loss functions (\textit{i.e.} Contrastive, Triplet and N-pairs) have poor performance in strict FAR ranges (such as BLUFR and MegaFace); after integrated with SST, they obtain significant improvement on these benchmarks.

\setlength{\tabcolsep}{1pt}
\begin{table}[ht]
\begin{center}
\caption{
Comparison of Semi-Siamese Training and the conventional training on deep data. In MegaFace, ``Id.'' refers to face identification rank1 accuracy with 1M distractors, and ``Veri.'' refers to face verification rate at 1e-6 FAR.}
\begin{tabular}{|c|p{0.9cm}<{\centering}|p{1.1cm}<{\centering}|p{0.9cm}<{\centering}|p{1.25cm}<{\centering}|p{1.2cm}<{\centering}|p{1.2cm}<{\centering}|p{1cm}<{\centering}|p{1cm}<{\centering}|}
\hline
\multirow{2}{*}{Method}&
\multirow{2}{*}{LFW}&
\multirow{2}{*}{AgeDB}&
\multirow{2}{*}{CFP}&
\multirow{2}{*}{CALFW}&
\multirow{2}{*}{CPLFW}&
\multirow{2}{*}{BLUFR}&
\multicolumn{2}{c|}{MegaFace}
\\ \cline{8-9}&&&&&&& Id. &Veri.\\
\hline\hline
DP-softmax&99.63&95.68&91.74&93.03&83.88&92.37&89.27&90.94\\
Contrastive&99.50&92.91&91.76&87.56&80.13&74.72&60.14&63.59\\
Triplet&99.47&93.32&94.49&89.32&82.25&79.10&65.65&69.18\\

N-pairs&99.53&94.58&93.43&92.10&83.15&85.19&76.87&78.28\\
\hline

SST (DP-softmax)&\textbf{99.68}&\textbf{96.24}&94.56&93.78&\textbf{86.04}&\textbf{94.78}&\textbf{92.08}&\textbf{93.57}\\
SST (Contrastive)&99.56&93.14&92.71&92.13&81.78&87.95&77.59&82.44\\
SST (Triplet)&99.50&94.30&93.30&92.05&82.67&89.72&81.76&83.29\\
SST (N-pairs)&99.65&96.12&\textbf{94.86}&\textbf{94.32}&84.74&94.17&91.72&93.48\\

\hline
\end{tabular}
\label{results_on_deep_data_s}
\end{center}
\end{table}

\subsection{Ablation Study}
In ablation study (the Section 4.2 of the main paper), we can see all the combination of “gallery queue” and “semi-siamese” (whatever network constraint or momentum) leads to the most significant boost for each training loss function. Besides, AM-softmax gains larger benefit than Arc-softmax from SST. We assume that the angular margin by Arc-softmax provides stronger supervision than AM-softmax, and such strong supervision distorts the feature space to some extent because the margin penalty performs on feature-feature pairs instead of features-FC pairs (especially training from scratch).

\subsection{Pretrain and Finetune}
In pretrain and finetune experiment (the Section 4.6 of the main paper), we can find the different improvement for softmax-based methods (softmax, AM-softmax, Arc-softmax) and pair/triplet-based methods (contrastive, triplet, N-pairs). We argue the heavy parameters of original softmax-based methods in classification FC layer brings the sub-optimal results in this experiment. After integrating with SST, the FC layer is replaced by an updating feature queue, which can significantly alleviate the optimization issue. Meanwhile, the pair/triplet-based methods adopt features rather than FC layer in the original version. So, the benefit brought by SST for softmax-based methods is larger than for pair/triplet-based methods in the finetuning stage.

\clearpage

\bibliographystyle{splncs04}

\begin{thebibliography}{10}
\providecommand{\url}[1]{\texttt{#1}}
\providecommand{\urlprefix}{URL }
\providecommand{\doi}[1]{https://doi.org/#1}

\bibitem{cao2018vggface2}
Cao, Q., Shen, L., Xie, W., Parkhi, O.M., Zisserman, A.: Vggface2: A dataset for recognising faces across pose and age. In: 2018 13th IEEE International Conference on Automatic Face \& Gesture Recognition (FG 2018). pp. 67--74 e(2018)

\bibitem{chen2018mobilefacenets}
Chen, S., Liu, Y., Gao, X., Han, Z.: Mobilefacenets: Efficient cnns for accurate real-time face verification on mobile devices. In: Chinese Conference on Biometric Recognition. pp. 428--438 (2018)

\bibitem{cheng2017know}
Cheng, Y., Zhao, J., Wang, Z., Xu, Y., Jayashree, K., Shen, S., Feng, J.: Know you at one glance: A compact vector representation for low-shot learning. In: Proceedings of the IEEE International Conference on Computer Vision Workshops. pp. 1924--1932 (2017)

\bibitem{cheng2018surveillance}
Cheng, Z., Zhu, X., Gong, S.: Surveillance face recognition challenge. arXiv preprint arXiv:1804.09691  (2018)

\bibitem{choe2017face}
Choe, J., Park, S., Kim, K., Hyun~Park, J., Kim, D., Shim, H.: Face generation for low-shot learning using generative adversarial networks. In: Proceedings of the IEEE International Conference on Computer Vision Workshops. pp. 1940--1948 (2017)

\bibitem{chopra2005learning}
Chopra, S., Hadsell, R., LeCun, Y.: Learning a similarity metric discriminatively, with application to face verification. In: IEEE Computer Society Conference on Computer Vision and Pattern Recognition. vol.~1, pp. 539--546 (2005)

\bibitem{deng2019arcface}
Deng, J., Guo, J., Xue, N., Zafeiriou, S.: Arcface: Additive angular margin loss for deep face recognition. In: Proceedings of the IEEE Conference on Computer Vision and Pattern Recognition. pp. 4690--4699 (2019)

\bibitem{dosovitskiy2014discriminative}
Dosovitskiy, A., Springenberg, J.T., Riedmiller, M., Brox, T.: Discriminative unsupervised feature learning with convolutional neural networks. In: Advances in neural information processing systems. pp. 766--774 (2014)

\bibitem{fei2006one}
Fei-Fei, L., Fergus, R., Perona, P.: One-shot learning of object categories. IEEE transactions on pattern analysis and machine intelligence \textbf{28}(4),  594--611 (2006)

\bibitem{feng2018wing}
Feng, Z.H., Kittler, J., Awais, M., Huber, P., Wu, X.J.: Wing loss for robust facial landmark localisation with convolutional neural networks. In:Proceedings of the IEEE Conference on Computer Vision and Pattern Recognition. pp. 2235--2245 (2018)

\bibitem{guo2017one}
Guo, Y., Zhang, L.: One-shot face recognition by promoting underrepresented classes. arXiv preprint arXiv:1707.05574  (2017)

\bibitem{guo2016ms}
Guo, Y., Zhang, L., Hu, Y., He, X., Gao, J.: Ms-celeb-1m: A dataset and benchmark for large-scale face recognition. In: European Conference on Computer Vision. pp. 87--102 (2016)

\bibitem{hadsell2006dimensionality}
Hadsell, R., Chopra, S., LeCun, Y.: Dimensionality reduction by learning an invariant mapping. In: IEEE Computer Society Conference on Computer Vision and Pattern Recognition. vol.~2, pp. 1735--1742 (2006)

\bibitem{he2019momentum}
He, K., Fan, H., Wu, Y., Xie, S., Girshick, R.: Momentum contrast for unsupervised visual representation learning. In: Proceedings of the IEEE Conference on Computer Vision and Pattern Recognition (2020)

\bibitem{he2016deep}
He, K., Zhang, X., Ren, S., Sun, J.: Deep residual learning for image recognition. In: Proceedings of the IEEE conference on computer vision and pattern recognition. pp. 770--778 (2016)

\bibitem{hu2018squeeze}
Hu, J., Shen, L., Sun, G.: Squeeze-and-excitation networks. In: Proceedings of the IEEE conference on computer vision and pattern recognition. pp. 7132--7141 (2018)

\bibitem{huang2008labeled}
Huang, G.B., Mattar, M., Berg, T., Learned-Miller, E.: Labeled faces in the wild: A database forstudying face recognition in unconstrained environments (2008)

\bibitem{kemelmacher2016megaface}
Kemelmacher-Shlizerman, I., Seitz, S.M., Miller, D., Brossard, E.: The megaface benchmark: 1 million faces for recognition at scale. In: Proceedings of the IEEE Conference on Computer Vision and Pattern Recognition. pp. 4873--4882 (2016)

\bibitem{liao2014benchmark}
Liao, S., Lei, Z., Yi, D., Li, S.Z.: A benchmark study of large-scale unconstrained face recognition. In: IEEE international joint conference on biometrics. pp.~1--8 (2014)

\bibitem{liu2019adaptiveface}
Liu, H., Zhu, X., Lei, Z., Li, S.Z.: Adaptiveface: Adaptive margin and sampling for face recognition. In: Proceedings of the IEEE Conference on Computer Vision and Pattern Recognition. pp. 11947--11956 (2019)

\bibitem{liu2017sphereface}
Liu, W., Wen, Y., Yu, Z., Li, M., Raj, B., Song, L.: Sphereface: Deep hypersphere embedding for face recognition. In: Proceedings of the IEEE conference on computer vision and pattern recognition. pp. 212--220 (2017)

\bibitem{moschoglou2017agedb}
Moschoglou, S., Papaioannou, A., Sagonas, C., Deng, J., Kotsia, I., Zafeiriou, S.: Agedb: the first manually collected, in-the-wild age database. In: Proceedings of the IEEE Conference on Computer Vision and Pattern Recognition
  Workshops. pp. 51--59 (2017)


\bibitem{ranjan2017l2}
Ranjan, R., Castillo, C.D., Chellappa, R.: L2-constrained softmax loss for discriminative face verification. arXiv preprint arXiv:1703.09507  (2017)

\bibitem{schroff2015facenet}
Schroff, F., Kalenichenko, D., Philbin, J.: Facenet: A unified embedding for face recognition and clustering. In: Proceedings of the IEEE conference on computer vision and pattern recognition. pp. 815--823 (2015)

\bibitem{sengupta2016frontal}
Sengupta, S., Chen, J.C., Castillo, C., Patel, V.M., Chellappa, R., Jacobs, D.W.: Frontal to profile face verification in the wild. In: 2016 IEEE Winter Conference on Applications of Computer Vision. pp.~1--9 (2016)

\bibitem{Simonyan2014Very}
Simonyan, K., Zisserman, A.: Very deep convolutional networks for large-scale image recognition. CoRR  \textbf{abs/1409.1556} (2015)

\bibitem{sohn2016improved}
Sohn, K.: Improved deep metric learning with multi-class n-pair loss objective. In: Advances in Neural Information Processing Systems. pp. 1857--1865 (2016)

\bibitem{sun2014deep}
Sun, Y., Chen, Y., Wang, X., Tang, X.: Deep learning face representation by joint identification-verification. In: Advances in neural information processing systems. pp. 1988--1996 (2014)

\bibitem{taigman2014deepface}
Taigman, Y., Yang, M., Ranzato, M., Wolf, L.: Deepface: Closing the gap to human-level performance in face verification. In: Proceedings of the IEEE conference on computer vision and pattern recognition. pp. 1701--1708 (2014)

\bibitem{trillionpairs.org}
trillionpairs.org: Ms-Celeb-1M-v1c. \url{http://trillionpairs.deepglint.com/overview}

\bibitem{wang2017residual}
Wang, F., Jiang, M., Qian, C., Yang, S., Li, C., Zhang, H., Wang, X., Tang, X.: Residual attention network for image classification. In: Proceedings of the IEEE Conference on Computer Vision and Pattern Recognition. pp. 3156--3164 (2017)

\bibitem{wang2018additive}
Wang, F., Cheng, J., Liu, W., Liu, H.: Additive margin softmax for face verification. IEEE Signal Processing Letters  \textbf{25}(7),  926--930 (2018)


\bibitem{wang2018cosface}
Wang, H., Wang, Y., Zhou, Z., Ji, X., Gong, D., Zhou, J., Li, Z., Liu, W.: Cosface: Large margin cosine loss for deep face recognition. In: Proceedings of the IEEE Conference on Computer Vision and Pattern Recognition. pp. 5265--5274 (2018)

\bibitem{wang2018feature}
Wang, L., Li, Y., Wang, S.: Feature learning for one-shot face recognition. In: 2018 25th IEEE International Conference on Image Processing. pp. 2386--2390 (2018)

\bibitem{wang2019co}
Wang, X., Wang, S., Wang, J., Shi, H., Mei, T.: Co-mining: Deep face recognition with noisy labels. In: Proceedings of the IEEE International Conference on Computer Vision. pp. 9358--9367 (2019)

\bibitem{wang2019mis}
Wang, X., Zhang, S., Wang, S., Fu, T., Shi, H., Mei, T.: Mis-classified vector guided softmax loss for face recognition. In: Proceedings of the AAAI Conference on Artificial Intelligence (2020)

\bibitem{wen2016discriminative}
Wen, Y., Zhang, K., Li, Z., Qiao, Y.: A discriminative feature learning approach for deep face recognition. In: European conference on computer vision. pp. 499--515 (2016)

\bibitem{wu2017low}
Wu, Y., Liu, H., Fu, Y.: Low-shot face recognition with hybrid classifiers. In: Proceedings of the IEEE International Conference on Computer Vision Workshops. pp. 1933--1939 (2017)

\bibitem{wu2018unsupervised}
Wu, Z., Xiong, Y., Yu, S.X., Lin, D.: Unsupervised feature learning via non-parametric instance discrimination. In: Proceedings of the IEEE Conference on Computer Vision and Pattern Recognition. pp. 3733--3742 (2018)


\bibitem{yi2014learning}
Yi, D., Lei, Z., Liao, S., Li, S.Z.: Learning face representation from scratch. arXiv preprint arXiv:1411.7923  (2014)

\bibitem{yin2019feature}
Yin, X., Yu, X., Sohn, K., Liu, X., Chandraker, M.: Feature transfer learning for face recognition with under-represented data. In: Proceedings of the IEEE Conference on Computer Vision and Pattern Recognition. pp. 5704--5713 (2019)

\bibitem{zhang2017faceboxes}
Zhang, S., Zhu, X., Lei, Z., Shi, H., Wang, X., Li, S.Z.: Faceboxes: A cpu real-time face detector with high accuracy. In: 2017 IEEE International Joint Conference on Biometrics. pp.~1--9 (2017)

\bibitem{zhang2019adacos}
Zhang, X., Zhao, R., Qiao, Y., Wang, X., Li, H.: Adacos: Adaptively scaling cosine logits for effectively learning deep face representations. In: Proceedings of the IEEE Conference on Computer Vision and Pattern Recognition. pp. 10823--10832 (2019)

\bibitem{zhao2019regularface}
Zhao, K., Xu, J., Cheng, M.M.: Regularface: Deep face recognition via exclusive regularization. In: Proceedings of the IEEE Conference on Computer Vision and Pattern Recognition. pp. 1136--1144 (2019)

\bibitem{zheng2018cross}
Zheng, T., Deng, W.: Cross-pose lfw: A database for studying crosspose face recognition in unconstrained environments. Beijing University of Posts and Telecommunications, Tech. Rep pp. 18--01 (2018)

\bibitem{zheng2017cross}
Zheng, T., Deng, W., Hu, J.: Cross-age lfw: A database for studying cross-age face recognition in unconstrained environments. arXiv preprint arXiv:1708.08197  (2017)


\bibitem{zhu2019large}
Zhu, X., Liu, H., Lei, Z., Shi, H., Yang, F., Yi, D., Qi, G., Li, S.Z.: Large-scale bisample learning on id versus spot face recognition. International Journal of Computer Vision vol.127, pp. 684--700 (2019)

\bibitem{zhuang2019local}
Zhuang, C., Zhai, A.L., Yamins, D.: Local aggregation for unsupervised learning of visual embeddings. In: Proceedings of the IEEE International Conference on Computer Vision. pp. 6002--6012 (2019)

\end{thebibliography}

\end{document}